\documentclass[journal,pdf]{IEEEtran}

\ifCLASSINFOpdf
\usepackage[pdftex]{graphicx}
\else
\fi

\usepackage{cite}
\usepackage{url}
\usepackage{amsmath,amssymb}
\usepackage{bm}
\usepackage{bold-extra}
\usepackage{makecell}
\usepackage{multirow}
\usepackage{hyperref}

\graphicspath{ {./figs/} }

\begin{document}

\title{Attentive Deep Regression Networks for Real-Time Visual Face Tracking in 
Video Surveillance}

\author{Safa~Alver and~Ugur~Halici
\thanks{S.~Alver was with the Department of Electrical and Electronics 
Engineering, Middle East Technical University, Ankara 68000, Turkey. (e-mail: 
alver.safa@metu.edu.tr).}
\thanks{U.~Halici is with the Department of Electrical and Electronics 
Engineering, Department of Biomedical Engineering and Neuroscience \& 
Neurotechnology Graduate Program, Middle East Technical University, Ankara 
68000, Turkey. (e-mail: halici@metu.edu.tr).}}

\markboth{Submitted August 2019}{Alver \& Halici: Attentive Deep Regression 
Networks for Real-Time Visual Face Tracking in Video Surveillance}
\maketitle

\begin{abstract}
Visual face tracking is one of the most important tasks in video surveillance 
systems. However, due to the variations in pose, scale, expression, and 
illumination it is considered to be a difficult task. Recent studies show that 
deep learning methods have a significant potential in object tracking tasks and 
adaptive feature selection methods can boost their performance. Motivated by 
these, we propose an end-to-end attentive deep learning based tracker, that is 
build on top of the state-of-the-art GOTURN tracker, for the task of real-time 
visual face tracking in video surveillance. Our method outperforms the 
state-of-the-art GOTURN and IVT trackers by very large margins and it achieves 
speeds that are very far beyond the requirements of real-time tracking. 
Additionally, to overcome the scarce data problem in visual face tracking, we 
also provide bounding box annotations for the G1 and G2 sets of ChokePoint 
dataset and make it suitable for further studies in face tracking under 
surveillance conditions.
\end{abstract}


\IEEEpeerreviewmaketitle

\section{Introduction}

\IEEEPARstart{V}{ideo} surveillance systems are widely deployed in both public 
and private places for the purpose of verifying or recognizing the individuals 
of interest. However, before the high level recognition or verification tasks, 
the faces have to be detected and tracked. Thus face tracking is one of the 
crucial tasks in video surveillance systems. Due to the variations in pose, 
scale, expression, and illumination it is considered to be a difficult task by 
itself. 

While the current learning-based video surveillance face trackers as IVT 
\cite{Ross:2008:ILR:1345995.1346002}, TLD \cite{Kalal:2012:TRA:2225045.2225082} 
and DSCT \cite{Wang2012DSCT}, which are compared by Dewan \textit{et al.} 
\cite{dewan_2013}, perform up to certain degrees, they cannot run at speeds 
that are required for real-time tracking (above 25 FPS). Additionally, these 
learning-based methods cannot avoid the drifting problem caused by learning the 
background or occlusion. Although the color histogram-assisted face tracker 
HAKLT proposed by Lan \textit{et al.} \cite{Lan2016} can perform real-time 
tracking (above 100 FPS), it makes use of the semantically weak color 
information where any similarly colored distractor around the face can lead to 
tracking the wrong target. Importantly, none of the face trackers in these 
studies make use of the powerful hierarchical features that can serve very well 
in representing and thus tracking human faces under various harsh conditions.

Most recently, deep learning \cite{lecun2015deep, schmid_review} based 
approaches has yielded significant performance increase in a wide variety of 
computer vision tasks as image classification \cite{ 
Krizhevsky_imagenetclassification, Simonyan14c}, object detection 
\cite{Girshick2014RichFH}, semantic segmentation \cite{Hong2015DecoupledDN, 	
Shelhamer:2017:FCN:3069214.3069246} and face verification 
\cite{Taigman14l.:deepface:}. Such great successes of deep learning based 
approaches is attributed mostly to their generalization capability and 
representation power. Despite the difficulties and challenges present in face 
tracking, and object tracking in general, deep learning based approaches have 
also shown state-of-the-art results in the recent visual object tracking (VOT) 
challenges. Starting from 2015, the deep learning based trackers have performed 
among the top in the VOT challenges \cite{Kristan2015a, Kristan2016a, 
Kristan2017a,Kristan2018a}. Further, the ones that are trained in an 
offline manner have reached speeds that surpass the real-time tracking 
requirements.

These deep learning based approaches use powerful hierarchical features 
that serve well in representing targets. However, different features may have 
different effects in tracking different objects. Using all of the features is 
neither efficient nor effective. Because of this, several adaptive feature 
selection methods have been developed \cite{He2018ATS, Wang_2018_CVPR} and they 
have been shown to be useful in object tracking tasks.

Motivated by these, we propose an end-to-end attentive deep learning based 
tracker, that is built on top of the state-of-the-art GOTURN tracker 
\cite{held2016learning}, for the task of real-time visual face tracking under 
surveillance conditions. We choose the GOTURN tracker as our starting 
point due to its high performance and simple end-to-end form. Another very 
important property is its ability to run at 100 FPS, which makes it one 
of the fastest \cite{Krebs2017ASO} trackers. Evaluation results show that our 
proposed tracker outperforms the state-of-the-art GOTURN and IVT trackers by 
very large margins. Furthermore, it runs at speeds ($\sim$140 FPS) that are 
very far beyond the requirements of real-time tracking. The main contributions 
of this letter are as follows:

\begin{itemize}
	\item  We show that a deep learning based generic object tracker can be 
	trained to track faces under surveillance conditions. A thorough search of 
	the relevant literature had yielded no published study on using deep 
	learning methods for the task of visual face tracking under surveillance 
	conditions.
	
	\item We take the state-of-the-art, real-time, single-target GOTURN tracker 
	and improve it using three main extensions. Although we use this network 
	for face tracking, it can also be used for any real-time single-target 
	visual object tracking task without any further modification in the 
	architecture. It just needs to be retrained with the domain specific 
	dataset in an offline manner.
	
	\item We provide bounding box annotations for the G1 and G2 sets of the 
	ChokePoint dataset \cite{wong_cvprw_2011} and make it suitable for further 
	studies in visual face tracking under surveillance conditions. The original 
	dataset only has person IDs and eye locations, making it incompatible with 
	the task of visual tracking.
\end{itemize}

\section{Method}

\subsection{Network Architecture}

\begin{figure}
	\centering
	\includegraphics[scale=0.4]{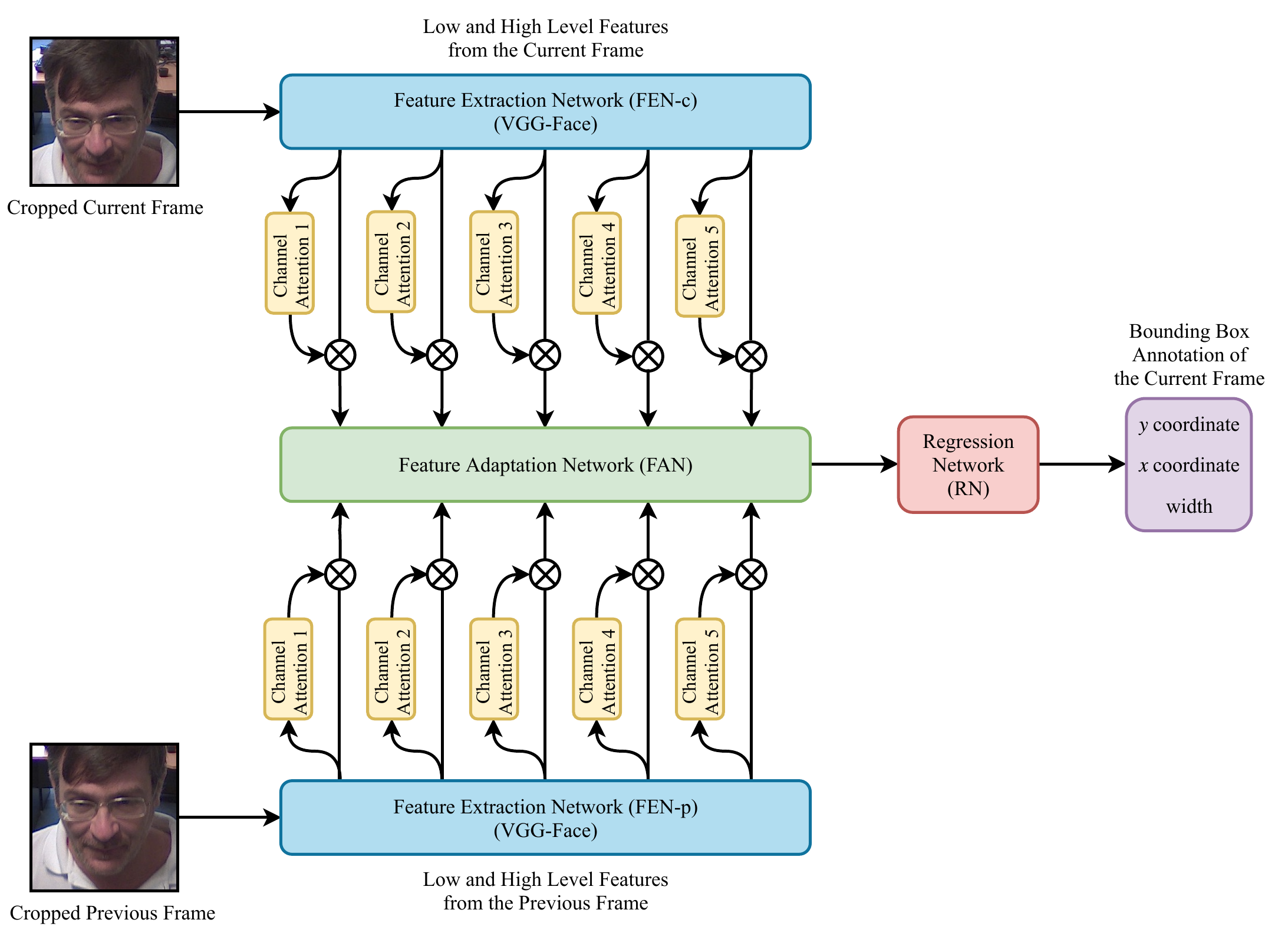}
	\caption{The network architecture of the proposed AFTN. After the features
		are extracted by the pretrained FENs, they are passed as input to the 
		CANs to get weighted. After weighting, the weighted features are 
		concatenated in the channel dimension via the FAN and then they are 
		passed to the RN for regressing the bounding box of the current frame.}
	\label{fig:network}
\end{figure}

We propose a face tracking network named Attentive Face Tracking Network (AFTN) 
whose architecture is as in Fig.\ \ref{fig:network}. The input to the network 
is a pair of 224$\times$224 RGB images that are cropped from the previous and 
current frames. These images act as the target object and the search region, 
respectively. The output is a tuple with three elements that describe the 
location and size of the target's bounding box within the 224$\times$224 
search region.

After choosing the GOTURN tracker \cite{held2016learning} as our starting 
point, we extend it using three main extensions. First is the usage of all
low and high level features in the tracking process. Making use of the lower 
level features has already been shown useful in the tracking literature 
\cite{Ma:2015:HCF:2919332.2919700, Wang2015VisualTW}. Second is the usage of 
a fusion network in the regression network which is adopted from the studies of 
Akkaya and Halici \cite{Akkaya_Thesis_16, Akkaya-Halici}. Third is 
the usage of a channel attention mechanism as in the case of He \textit{et 
al.} \cite{He2018ATS}. However, rather than using it for just the static 
channels of the last two layers, we use it to adaptively weight the dynamically 
changing channels from all of the layers. Readers who are interested in the 
detailed step-by-step building procedure of AFTN are referred to the MSc thesis 
\cite{Alver_Thesis_19} of Alver.

In detail, AFTN is composed of the following four parts:

\subsubsection{Feature Extraction Network (FEN)} 

The FENs are composed of convolutional (Conv) layers of the pretrained 
five-layer VGG-Face \cite{Parkhi15} network\footnote{Information on the 
architecture and the pretrained model in PyTorch can be reached at: 
\url{http://www.robots.ox.ac.uk/~albanie/pytorch-models.html}. We use the 
vgg-m-face model as it allows fast inference.} and they act as frozen feature 
extractors for the previous and current frames. While shallow layers of the 
FENs extract simple low-level features as edges and corners, the deeper layers 
extract more complex high-level features as the semantics of the target. 
Although these high-level features are very useful in tasks that require 
semantic information, their receptive fields in the input are very large, 
making them less precise in localizing targets. Therefore, we use both the 
high-level and low-level features together. Since it is not immediately 
clear which level features will serve well in tracking, we basically use 
all them and pass them as inputs to the channel attention networks for 
adaptive weighting.
	
\subsubsection{Channel Attention Network (CAN)} 

The CANs are simple two-layer multilayer perceptrons (MLP) that are used for 
weighting the input channels. The first fully connected (FC) layer has 36 units 
with ReLU activations and the following FC layer has a single unit with a 0.5 
biased sigmoid activation corresponding to the weight coefficient for the input
channel. These coefficients are then used for weighting the channels according 
to their importance in tracking. Since we use 5 different layers from the FENs, 
there are also 5 different CANs. CANs are trained during the offline training 
process.
	
\subsubsection{Feature Adaptation Network (FAN)}

The FAN is used for concatenating all the weighted features in their channel 
dimension. It has no learnable parameters.
	
\subsubsection{Regression Network (RN)}

The RN is composed of a Conv layer followed by three FC layers and it is used 
for regressing the target's bounding box from the input concatenated features. 
The Conv layer is used as a fusion network for fusing the spatial and semantic 
information in the concatenated features. It has 256 1$\times$1 kernels with 
stride 1, zero-padding 0, batch norm with default parameters 
\cite{DBLP:journals/corr/IoffeS15} and ReLU activations. The first two FC 
layers have 4096 units with ReLU activations and 0.5 dropout 
\cite{JMLR:v15:srivastava14a}. The last FC layer has 3 units with no 
activations corresponding to the bounding box annotation. As CANs, the RN is 
also trained during the offline training process.

\subsection{Channel Attention Mechanism}

\begin{figure}
	\centering
	\includegraphics[scale=0.45]{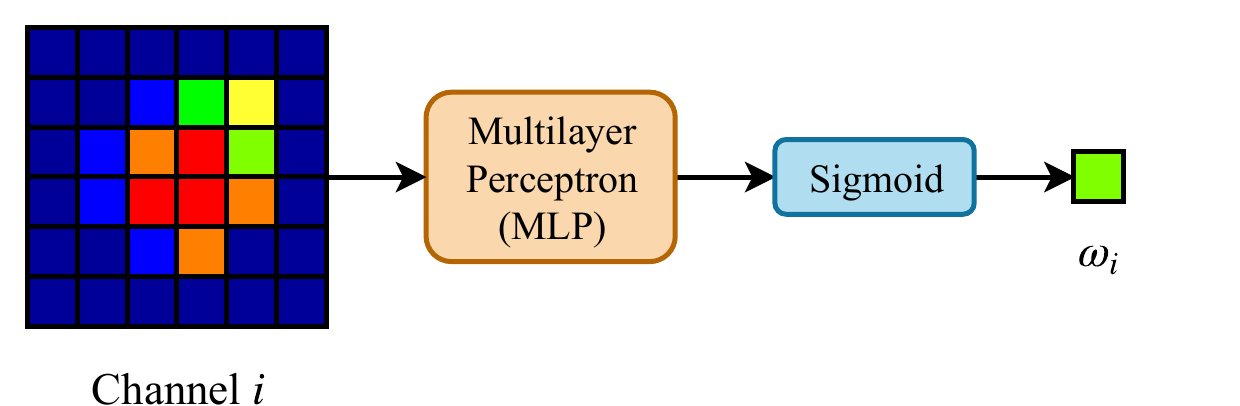}
	\caption{The channel attention mechanism that generates the weighting 
	coefficient $\omega_i$ for the $i$-th channel.}
	\label{fig:att}
\end{figure}

For the task of tracking, some channels in the FENs may be unnecessary or even 
harmful, whereas some may be very useful. In order to automate this channel 
selection process, we use a channel attention mechanism with details in 
Fig.\ \ref{fig:att}. Specifically, we first flatten the 6$\times$6 channels and 
then pass them through the CANs with biased sigmoid outputs to obtain their 
weight coefficients. Sigmoids have a bias to ensure that no channel is 
suppressed down to zero. Since the lower layers in the FENs have channels with 
sizes 54$\times$54 and 13$\times$13, we first max-pool them to match the 
6$\times$6 size of the last layer and then pass them through the CANs. In more 
detail, for the 54$\times$54 channels, we use two cascaded pooling layers with 
kernel sizes 6$\times$6 and 3$\times$3 and strides 4 and 2. And for the rest of 
the 13$\times$13 channels, we use a pooling layer with kernel size 3$\times$3 
and stride 2. After the weighting coefficients are obtained, the activations in 
the channels are multiplied with their corresponding weights. By this way, 
while the unnecessary features will get suppressed, the useful ones will get 
pushed up.

\subsection{Dataset}

In order to test our proposed face tracker, we use the G1 and G2 sets of 
the ChokePoint dataset \cite{wong_cvprw_2011}. This dataset contains 30 FPS 
sequences with frames of size 800$\times$600. However, the annotations provided 
with it are only the person IDs and eye locations that are not compatible 
with the task of visual tracking. Thus, we annotate\footnote{We perform the 
annotation by first using a face detector \cite{dlib09} to detect the faces and 
then we manually go over the bounding boxes to correct for the mistakes that 
the detector makes. The bounding box annotations can be reached at: 
\url{https://github.com/alversafa/chokepoint-bbs}} the frames in the G1 and 
G2 sets with bounding boxes and make the ChokePoint dataset suitable for visual 
face tracking. This newly formed dataset consists of 432 different video 
sequences (216 in G1 and 216 in G2) each having only a single person present at 
a given time and 37,307 frames (16,665 in G1 and 20,652 in G2) with a face. The 
average length of a sequence is 95.6 frames for G1 and 77.1 frames for G2. For 
the evaluation, we use the baseline verification protocol for the ChokePoint 
dataset, where we first use G1 to train our network and use G2 to evaluate it 
and then do the reverse. In the end, we report the average performances of our 
trackers on the evaluation sets.

\subsection{Offline Training and Online Tracking}

In the offline training phase, we randomly choose pairs of successive frames 
from our training dataset with a batch size of 50. We then crop these pair of 
frames using twice the size of the previous frame's bounding box and resize 
them to 224$\times$224 (input size of VGG-Face). We also subtract the mean of 
the dataset that was used in training VGG-Face. Since we feed the cropped 
images to the network, we also transform the bounding box annotations of the 
800$\times$600 images to the 224$\times$224 ones and use these transformed 
annotations in the training process. After these preprocessings, we feed the 
data batch to the network and compute the L1 loss between the ground-truth 
annotations and the predicted ones. We then backpropagate from this loss and 
use the Adam optimizer \cite{DBLP:journals/corr/KingmaB14} with a constant 
learning rate of $1e-5$. The training is done for 10 epochs and PyTorch 
\cite{paszke2017automatic} is used for implementation. L2 regularization is 
applied to the weights with a penalty factor of $1e-3$. Importantly, the 
network is trained in a fully end-to-end manner.

In the online tracking phase, first, the initial two frames are read from the 
sequence and the same preprocessing steps in the offline training phase are 
applied. Then, both of the frames are fed to the network to obtain the bounding 
box of the current frame within the 224$\times$224 search region. This bounding 
box is then transformed back to its corresponding location in the 
800$\times$600 frame. After this, the current frame is used as the previous 
frame and a new frame is read as the current frame. Using the predicted 
bounding box annotation from the previous step, the consecutive frames are 
again cropped-resized and fed to the network for regressing the bounding box in 
the newly read frame. This crop-resize-feed-read cycle continues for the rest 
of the frames and by this way tracking is performed.

\subsection{Evaluation Metrics}

In order to compare the performance of trackers, we use the complementary 
accuracy and robustness measures that were proposed by \u{C}ehovin \textit{et 
al.} \cite{Cehovin_Zajc2014a, Cehovin2016VisualOT}. However, we use slightly 
different metrics to account for the accuracy and robustness. Specifically, 
rather than the average overlap, we use the equivalent\footnote{See the 
supplementary material of \u{C}ehovin \textit{et al.} \cite{Cehovin_Zajc2014a, 
Cehovin2016VisualOT} for the proof.} area under the curve of the True Positive 
vs.\ Region Overlap Threshold (TP vs.\ ROT) plot to account for accuracy. And 
rather than the failure rate with a single threshold of 0, we use the area 
above the curve of the Failure Rate vs.\ Reinitialization Threshold (FR vs.\ 
RT) plot to account for robustness. This can be seen as an average of the 
failure rates for different RTs. These plots have the advantage of displaying 
the tracker's performances for not only one, but for thresholds ranging from 0 
to 1. In the end, we summarize a tracker's accuracy and robustness by using the 
overall score which is just the average of the two. For the speed comparison, 
we use the FPS values of the trackers.

The TP vs.\ ROT plot for a sequence is obtained by running a tracker over the 
sequence and calculating the ratio of frames where the region overlap between 
the ground-truth and tracker's bounding box is greater than the ROT. In this 
plot, the tracker gets reinitialized only if the region overlap becomes 0 (RT 
is 0). Similarly, the FR vs.\ RT plot is obtained by again running a tracker 
over the sequence and calculating the ratio of frames where the region overlap 
fails below the RT. In this plot, the tracker gets reinitialized if the region 
overlap falls below the RT.

\section{Results}
\label{chp:ch5}

\begin{figure}
	\centering
	\includegraphics[scale=0.39]{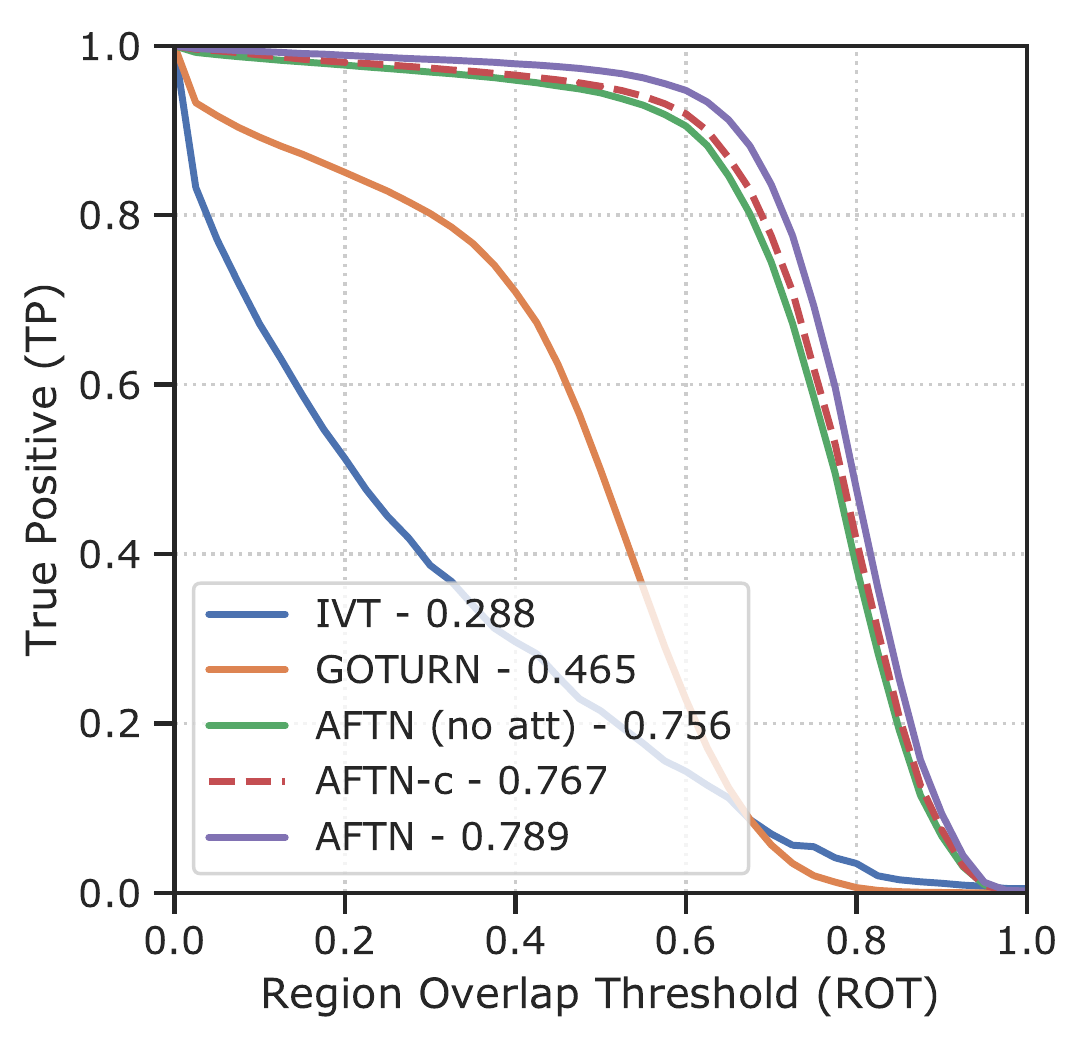}
	\includegraphics[scale=0.39]{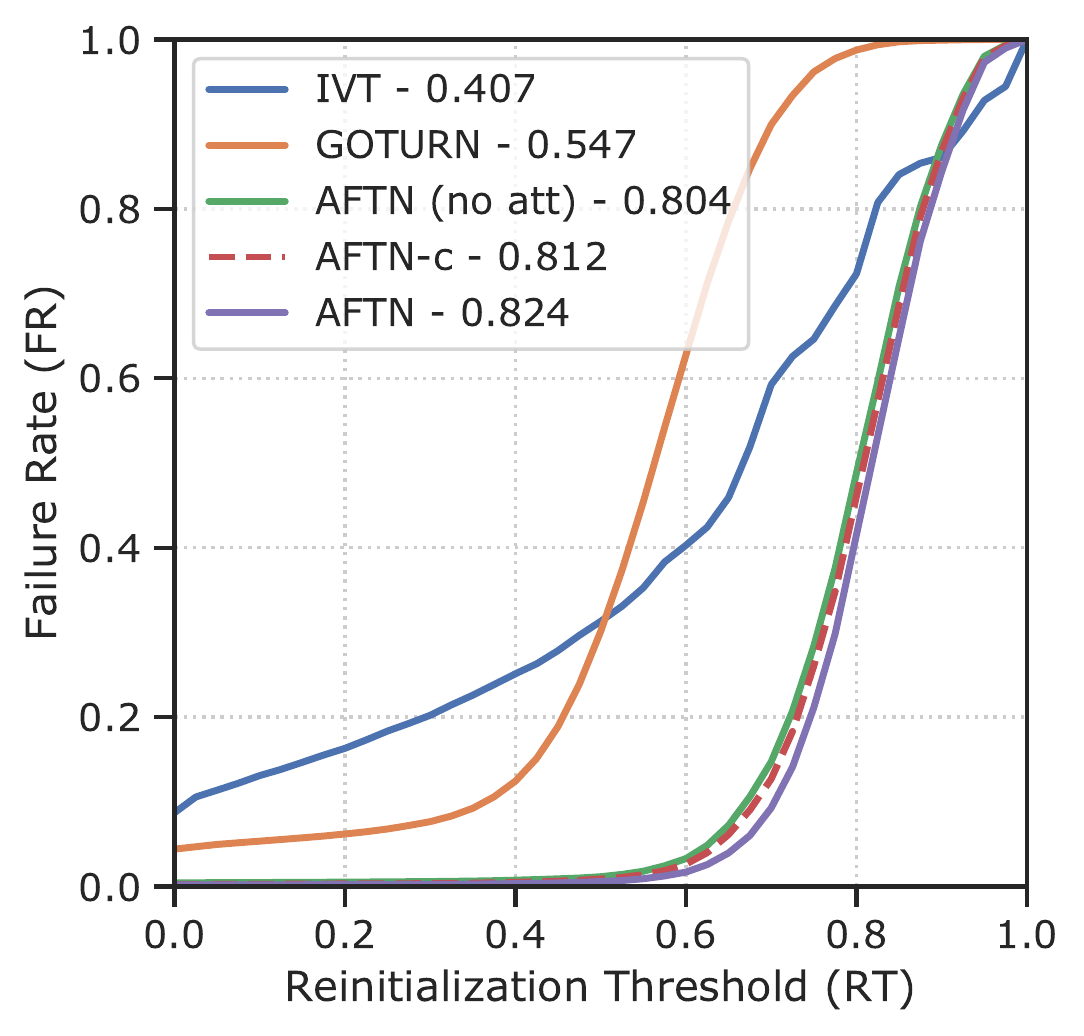}
	\caption{The TP vs.\ ROT (left) and FR vs.\ RT (right) plots for the IVT, 
	GOTURN, AFTN (no att), AFTN-c and AFTN trackers. The accuracy and 
	robustness values of the trackers are given next to their name in the 
	legend.}
	\label{fig:plots}
\end{figure}

The accuracy and robustness values of our proposed tracker (AFTN) are presented 
in Fig.\ \ref{fig:plots} and Table \ref{tab:overall}. For comparison, we also 
provide values of the IVT \cite{Ross:2008:ILR:1345995.1346002} (using the best 
hyperparameters proposed in the study and forcing it to output squares for 
fair comparison) and GOTURN \cite{held2016learning} (using VGG-Face 
\cite{Parkhi15} in place of the AlexNet \cite{Krizhevsky:2012} and again 
forcing it to output squares for fair comparison) trackers. Among the 
other surveillance trackers available in the literature, we do not perform 
comparisons with TLD \cite{Kalal:2012:TRA:2225045.2225082} and DSCT 
\cite{Wang2012DSCT} trackers as Dewan \textit{et al.} \cite{dewan_2013} have 
already shown that IVT performs better than the two. Finally, to examine 
the effect of the attention mechanism and the necessity of using of the 
previous frame, we also provide values of AFTN (no att) and AFTN-c, which are 
the versions of AFTN with no attention and no usage of the previous frame, 
respectively.

\begin{table}
	\caption{Accuracy, robustness, overall scores and speeds of the trackers}
	\begin{center}
		\begin{tabular}{ c | c | c | c | c}
			\Xhline{3\arrayrulewidth}
			\textbf{Tracker} & \textbf{Accuracy} & \textbf{Robustness} & 
			\textbf{Overall} & \textbf{Speed} \\ 
			\Xhline{3\arrayrulewidth}
			IVT & 0.288$\pm$0.054 & 0.407$\pm$0.063 & 0.348 & 12.9 \\ \hline
			GOTURN & 0.465$\pm$0.107 & 0.547$\pm$0.065 & 0.506 & 118.6 \\ \hline
			AFTN (no att) & 0.756$\pm$0.075 & 0.804$\pm$0.038 & 0.780 & 148.9 
			\\ \hline
			\textbf{AFTN-c} & \textbf{0.767}$\pm$0.079 & 
			\textbf{0.812}$\pm$0.036 & \textbf{0.790} & \textbf{183.4} \\ \hline
			\textbf{AFTN} & \textbf{0.789}$\pm$0.059 & 
			\textbf{0.824}$\pm$0.032 & \textbf{0.807} & \textbf{142.9} \\ \hline
		\end{tabular}
	\end{center}
	\label{tab:overall}
\end{table}

We see that the proposed AFTN outperforms both IVT and GOTURN by very large 
margins (overall score of 0.348, 0.506, and 0.807 for IVT, GOTURN, and AFTN, 
respectively). The IVT tracker performs even worse than the GOTURN tracker. 
This is expected as traditional trackers like IVT do not make use of 
the powerful hierarchical features present in deep learning based trackers. We 
also see that taking away the attention mechanism from AFTN results in a 
significant drop in the overall score (from 0.807 to 0.780). Lastly, we see 
that although not using the previous frame as input can cause a drop in the 
overall score (from 0.807 to 0.790), it can bring a significant jump in the 
speed (from 142.9 FPS to 183.4 FPS).

\begin{figure}
	\centering
	\includegraphics[scale=0.4]{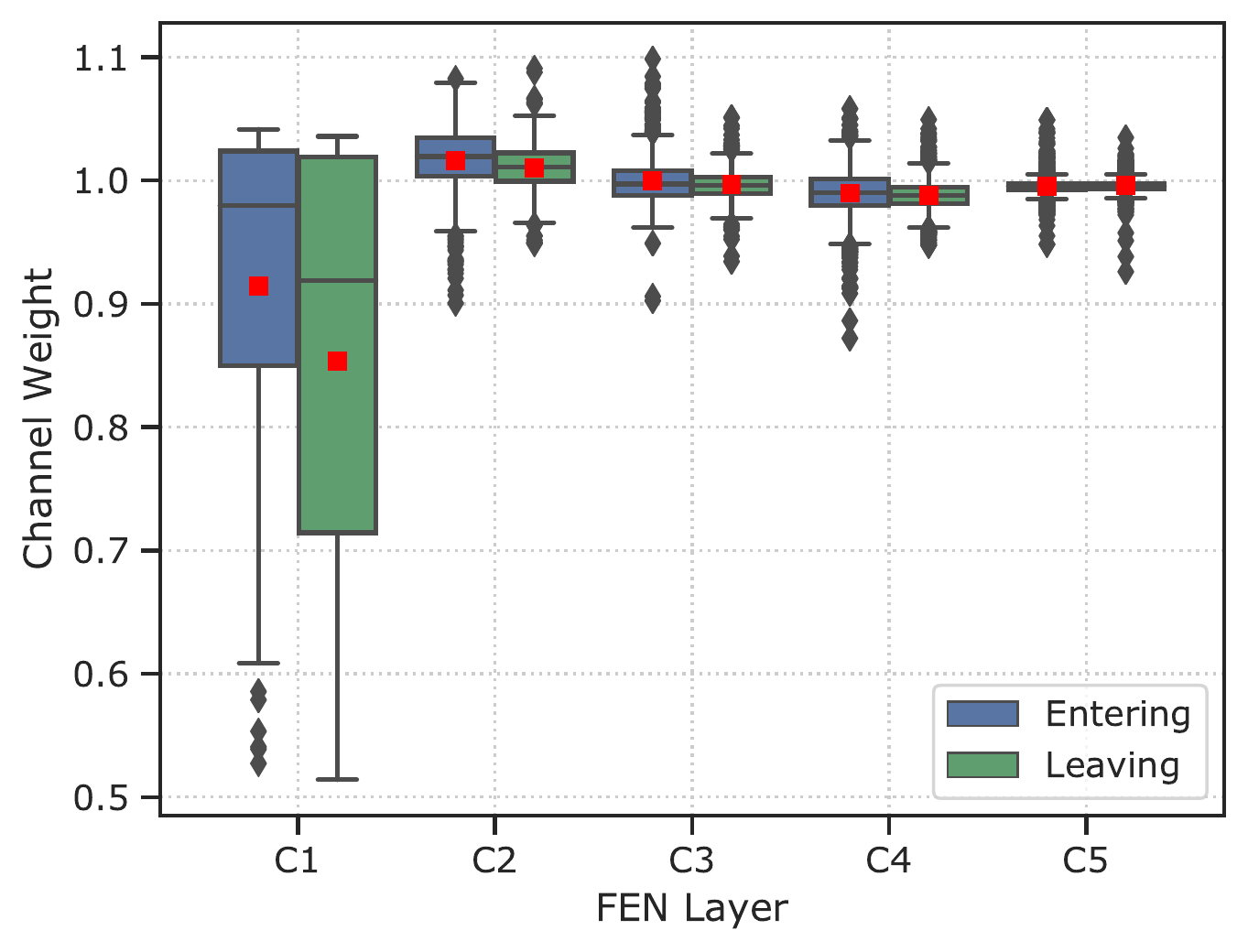}
	\includegraphics[scale=0.24]{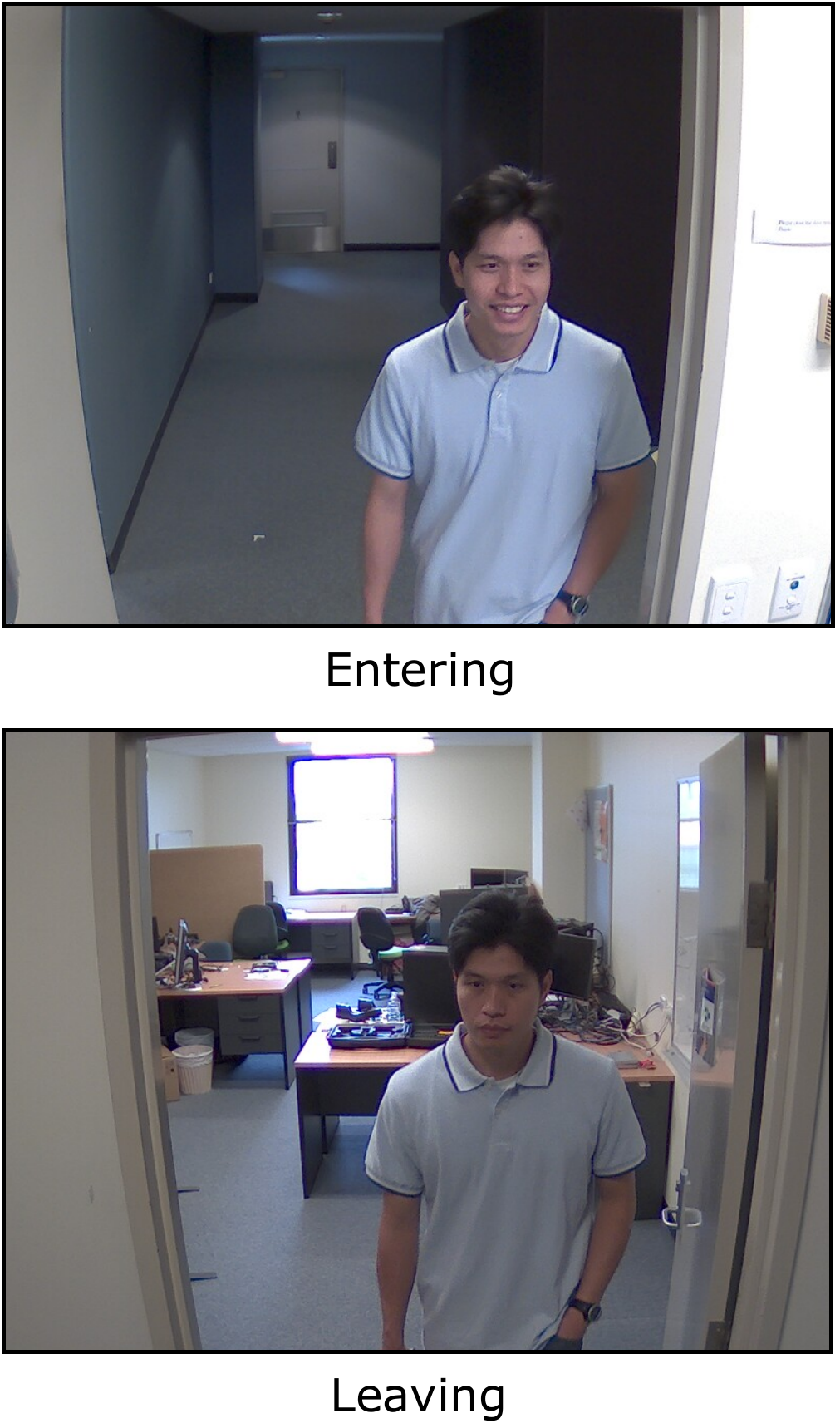}
	\caption{The average channel weights (left) of the AFTN in the entering and 
	leaving scenarios of person with ID 18 (right). CX corresponds to the Xth 
	Conv layer in the FENs. Red dots are the means.}
	\label{fig:att_results}
\end{figure}

In order to show the effect of the channel attention mechanism, in Fig. 
\ref{fig:att_results} we provide detailed information on the average channel 
weights of AFTN for the entering and leaving scenarios of person with ID 
18.\footnote{These scenarios are chosen from the P1E\_S4\_C1 and P1L\_S4\_C1 
sequences, respectively.} First, we see that C1 features get suppressed more 
compared to the other levels. This is expected as lower level features may 
contain generic information that is not useful for tracking. We also see that 
more features from C1 are suppressed in the leaving scenario compared to the 
entering one (with means of 0.85 and 0.92, respectively). This is also expected 
as the leaving scenario contains more distracting office objects in the 
background and the lower level features in C1 contain information on them. It 
should be noted that the weights for other persons and sequences also show a 
similar distribution.

In Table \ref{tab:overall}, we also provide the speeds of the trackers 
with unoptimized codes. It is clear that all of the deep learning based 
trackers can run at speeds that are very far beyond the 25 FPS requirement for 
real-time tracking. This is mainly due to the following two aspects: the 
trackers are trained fully offline with no online updating involved and only a 
single forward pass is enough for inferring the bounding box annotations. The 
usage of GPUs, rather than CPUs, is another important aspect that significantly 
contributes to this. It should be noted that in order to make a fair speed 
comparison, we ran all trackers on a machine equipped with an Intel Core 
i7-4790K 8 Core 4.00 GHz CPU and a single NVIDIA GeForce GTX Titan X GPU. We 
also enabled the benchmark mode of PyTorch during tracking.

From Table \ref{tab:overall} it is clear that the best performing trackers are 
AFTN and AFTN-c. Among these two, the former one has a higher overall score 
whereas the latter performs much faster. So which one of these two trackers 
should be used? If speed is a concern, then AFTN-c can be used as its overall 
score is comparable to AFTN. However, if the overall score is the major 
concern, then AFTN should be preferred as it has the highest overall score. So 
the choice of which tracker to use depends on the objective to maximize.

\section{Conclusion}

In this letter, we have proposed an attentive deep regression network, an 
extended version of the GOTURN tracker, for the task of real-time visual face 
tracking in video surveillance. Experimental results demonstrated that our 
proposed tracker outperforms the state-of-the-art GOTURN and IVT trackers by
very large margins and it achieves speeds ($\sim$140 FPS) that are very far 
beyond the requirements of real-time tracking. Further, we demonstrated the 
usefulness of the attention mechanism by showing that not using it can result 
in a significant drop in the overall score. We also ran experiments to check 
the necessity of using the previous frame as input and showed that it might not 
be necessary if speed is the major concern. Additionally, we provided bounding 
box annotations for the G1 and G2 sets of the ChokePoint dataset and made it 
suitable for further studies in surveillance face tracking. As a final comment, 
we highlight the need for more empirical benchmarking studies in (surveillance) 
face tracking for the rapid advancement of the field.

\bibliographystyle{IEEEtran}
\bibliography{IEEEabrv, alver_attentive_2019}

\end{document}